\pdfoutput=1
\documentclass[lettersize,journal]{IEEEtran}
\usepackage{amsmath,amsfonts,amssymb}
\usepackage{algorithmic}
\usepackage{algorithm}
\usepackage{array}
\usepackage{tabularx}
\usepackage{booktabs}
\usepackage[caption=false,font=normalsize,labelfont=sf,textfont=sf]{subfig}
\usepackage{textcomp}
\usepackage{stfloats}
\usepackage{url}
\usepackage{verbatim}
\usepackage{graphicx}
\usepackage{cite}
\usepackage{xcolor}
\usepackage{pifont}
\usepackage{rotating}
\usepackage{tabularray}
\usepackage{makecell}
\usepackage{siunitx}
\usepackage[hidelinks]{hyperref}

\newcommand{\genmark}{\ensuremath{\checkmark\mkern-6mu\raisebox{0.25ex}{\ensuremath{\scriptstyle\circ}}}}
\newcommand{\starmark}{\ensuremath{\checkmark\mkern-7mu\raisebox{0.35ex}{\ensuremath{\scriptstyle\ast}}}}

\begin{document}

\title{DGHMesh: A Large-scale Dual-radar mmWave Dataset and Generalization-focused Benchmark for Human Mesh Reconstruction}

\author{Rongxiao Guo, Qingchao Chen~\textsuperscript{\ding{41}}

\thanks{Rongxiao Guo is with the Institute of Medical Technology, Peking University Health Science Center, and the National Institute of Health Data Science at Peking University, and also with the State Key Laboratory of General Artificial Intelligence, Peking University, Beijing, China.}
\thanks{Qingchao Chen is with the National Institute of Health Data Science at Peking University and the State Key Laboratory of General Artificial Intelligence, Peking University, and also with the Institute of Medical Technology, Peking University Health Science Center, Beijing, China.}
\thanks{This work was supported by grants from the National Natural Science Foundation of China (62571009).}
\thanks{\ding{41} Corresponding author: Qingchao Chen.}
}



\IEEEpubid{%
\makebox[\columnwidth][l]{\begin{minipage}[t]{\columnwidth}\footnotesize\raggedright
\vspace{-1.2ex}
\hrule height 0.4pt\relax\vspace{0.75ex}
This work has been submitted to the IEEE for possible publication. Copyright may be transferred without notice, after which this version may no longer be accessible.
\end{minipage}}%
\hspace{\columnsep}%
\makebox[\columnwidth]{}
}\IEEEpubidadjcol

\maketitle

\begin{abstract}
Millimeter-wave (mmWave) radar has shown great potential for contactless, privacy-preserving, and robust human sensing, yet existing mmWave-based human mesh reconstruction (HMR) studies are still limited by the lack of benchmarks for generalization analysis under configuration shifts and fair comparison of different algorithms. To address the limitation, we present \textbf{DGHMesh}, a large-scale dual-radar mmWave dataset and generalization-focused benchmark for HMR. It contains data from 15 subjects performing 8 actions, with 360,000 synchronized frames collected from FMCW radar, SFCW radar, RGB images, and high-precision 3D HMR annotations. In addition, the dataset provides synchronized raw I/Q data from both radar modalities and accurately calibrated radar spatial positions. The benchmark is designed to evaluate HMR methods under diverse measurement configurations, including human position shifts, human orientation shifts, subarray size variations, and cross-subject settings. Based on DGHMesh, we also propose \textbf{mmPTM}, a query-based multi-radar fusion framework that jointly exploits point clouds and imaging tubes for HMR. Extensive experiments are conducted against representative baselines under different settings. The results demonstrate that mmPTM consistently achieves outstanding accuracy and competitive generalization capability across multiple sub-benchmarks, validating the effectiveness of multi-radar fusion and the practical value of the proposed dataset and benchmark for mmWave-based HMR research. DGHMesh and mmPTM are publicly available at 
\url{https://github.com/SPIresearch/DGHMesh}.
(The complete benchmark and code will be released after paper publication)
\end{abstract}


\section{Introduction}

Millimeter-wave (mmWave) radar has attracted rapidly growing interest in the intelligent sensing community and has become an active research topic in recent years\cite{soni2025millimeter,kong2024survey}. It has been widely explored in a broad range of applications, including autonomous driving\cite{fan20244d}, target detection\cite{chen2023non}, gesture recognition\cite{li2022towards}, human activity recognition\cite{ding2025omnidirectional}, smart home systems\cite{zeng2025mmmotion}, and vital sign monitoring\cite{xiong2022vital}. Compared with conventional vision-based sensors, mmWave radars do not rely on visible-light imaging and can therefore operate reliably under low illumination, severe occlusion, and privacy-sensitive situations, making them highly attractive in practical applications\cite{wu2023non,sesyuk2024radar,xu2024radar}.

Among the various mmWave-based human sensing applications, healthcare monitoring is one of the most important and practically valuable directions\cite{zhang2025comprehensive}. Representative research topics include fall detection\cite{zhao2025mm}, sleep monitoring\cite{adhikari2024misleep}, respiration and heartbeat monitoring\cite{qin2026motion,wu2025non,huang2025concurrent}, and early disease screening\cite{das2022novel,zhao2024mmarrhythmia}. These studies are significant not only for home health management, but also for clinical assistance and long-term non-invasive physiological monitoring\cite{zhang2025comprehensive,review,wu2022health}. As this field continues to evolve, mmWave-based human sensing has moved beyond coarse activity recognition toward more fine-grained human understanding, where 3D human motion analysis and reconstruction have emerged as key directions\cite{chen2024toward}. 

In 3D human perception, existing approaches are generally divided into two categories: Human Pose Estimation (HPE) and Human Mesh Reconstruction (HMR)\cite{m4human}. HPE focuses on recovering body keypoints and skeletal structure\cite{hpe1}, whereas HMR aims to reconstruct the full 3D human mesh, thereby providing richer geometric information and more detailed body-shape representation\cite{hmr1}. Compared with HPE, HMR can characterize not only body pose, but also body shape, body-size variation, and local surface structure\cite{hmr2}. As a result, HMR is more informative for applications such as virtual reality, motion analysis, and digital human modeling\cite{hmr3}. However, mmWave-based HMR research is still in its early stage, and existing studies remain limited in terms of sensing modality, annotation accuracy, experimental setting diversity, and public dataset availability\cite{mmmesh,mmbody}.

From a theoretical perspective, mmWave-based HMR can be viewed as an ill-posed inverse problem: the objective is to reconstruct a dense 3D human mesh from sparse, noisy, and partially observable radar measurements. In the classical Hadamard sense, an inverse problem is considered well-posed only if a unique solution exists and depends continuously on the input data; otherwise, it is ill-posed, and a notorious issue is the sensitivity of the solution to small perturbations in the input data\cite{inverse1,inverse2}. In mmWave-based HMR, such perturbations are not limited to sensor noise, but are also induced by changes in sensing configuration, which alter the effective input distribution and directly influence the generalization capability of the reconstruction model. In particular, variations in human position, human orientation, subarray size, and subject identity can all lead to substantial shifts in the radar observations, thereby making the inverse mapping more ambiguous and less stable under practical sensing conditions.

\begin{table*}[t]
\centering
\scriptsize
\caption{Comparison of mmWave radar human sensing datasets and benchmarks}
\label{tab:dataset_comparison}

\begin{tblr}{
  cells = {c, m},
  colsep = 2pt,
  cell{1}{1} = {r=2}{},
  cell{1}{2} = {r=2}{},
  cell{1}{3} = {c=5}{},
  cell{1}{8} = {c=3}{},
  cell{1}{11} = {c=4}{},
  cell{1}{15} = {c=3}{},
  cell{3}{1} = {r=8}{},
  cell{11}{1} = {r=4}{},
  cell{14}{2} = {font=\bfseries},
  vline{3,8,11,15} = {1-Z}{0.03em},
  hline{1,Z} = {-}{0.1em},
  hline{3,11} = {-}{0.03em},
}
Task             & Dataset    & Output Modalities &                    &            &              &         & Annotations            &          &         & Statistics &        &        &          & Sub-benchmark       &                       &                     \\
                 &            & RGB               & \makecell[c]{Multi-mmWave\\Radar} & \makecell[c]{Point\\Cloud} & \makecell[c]{Radar\\Tensor} & \makecell[c]{Raw\\I/Q} & \makecell[c]{Radar Spatial\\Position} & Skeleton & 3D Mesh & \# Sub.       & \# Act. & \# Seq. & \# Frame & \makecell[c]{Different\\Positions} & \makecell[c]{Different\\Orientations} & \makecell[c]{Different\\Subarrays} \\
\makecell[c]{mmWave-\\based\\HPE} & RF-Pose3D†\cite{rfpose3d} & $\checkmark$                 &                    &            & $\checkmark$            &         &                        & $\starmark$       &         & $>$5           & 5      & -      & -        &                     &                       &                     \\
                 & mmPose†\cite{mmpose}    &                   &                    & $\checkmark$          &              &         &                        & $\starmark$       &         & 2            & 4      & -      & 40K      &                     &                       &                     \\
                 & Hupr\cite{hupr}       &                   &                    &            & $\checkmark$            &         &                        & $\starmark$       &         & 6            & 3      & 235    & 141K     &                     &                       &                     \\
                 & MARS\cite{mars}       &                   &                    & $\checkmark$          &              &         &                        & $\starmark$       &         & 4            & 10     & 80     & 40K      &                     &                       &                     \\
                 & mRI\cite{mri}        & $\checkmark$                 &                    & $\checkmark$          &              &         &                        & $\starmark$       &         & 20           & 12     & 300    & 160K     &                     &                       &                     \\
                 & mm-Fi\cite{mmfi}      & $\checkmark$                 &                    & $\checkmark$          &              &         &                        & $\starmark$       &         & 40           & 27     & 1080   & 321K     &                     &                       &                     \\
                 & RT-Pose\cite{rtpose}    & $\checkmark$                 &                    & $\genmark$          & $\checkmark$            & $\checkmark$       &                        & $\starmark$       &         & 10           & 6      & 240    & 72K      &                     &                       &                     \\
                 & MM-DCDR\cite{mmdcdr}    & $\checkmark$                 & $\checkmark$                  & $\checkmark$          & $\checkmark$            &         &                        & $\starmark$       &         & 11           & 8      & -      & 352K     &                     &                       &                     \\
\makecell[c]{mmWave-\\based\\HMR} & mmMesh†\cite{mmmesh}    & $\checkmark$                 &                    & $\checkmark$          &              &         &                        & $\checkmark$        & $\checkmark$       & 20           & 8      & -      & 480K     &                     &                       &                     \\
                 & mmBody\cite{mmbody}     & $\checkmark$                 &                    & $\checkmark$          &              &         &                        & $\checkmark$        & $\checkmark$       & 20           & 100    & -      & $>$200K    &                     &                       &                     \\
                 & M4Human\cite{m4human}    & $\checkmark$                 &                    & $\checkmark$          & $\checkmark$            &         &                        & $\checkmark$        & $\checkmark$       & 20           & 50     & 999    & 661K     &                     &                       &                     \\
                 & DGHMesh (Ours)      & $\checkmark$                 & $\checkmark$                  & $\checkmark$          & $\genmark$            & $\checkmark$       & $\checkmark$                      & $\checkmark$        & $\checkmark$       & 15           & 8      & 360    & 360K     & $\checkmark$                   & $\checkmark$                     & $\checkmark$                   \\

\end{tblr}

\vspace{2pt}
\begin{minipage}{\linewidth}
\raggedright\scriptsize
\begin{tabularx}{\linewidth}{@{}l@{\quad}X@{}}
† & indicates the dataset is not publicly available.\\
$\genmark$ & indicates that the corresponding radar modality is not directly provided by the dataset, but can be generated from the released raw I/Q data.\\
$\starmark$ & indicates that the skeleton annotations are generated from camera images with a labeling algorithm rather than from a high-precision motion capture system.
\end{tabularx}
\end{minipage}

\end{table*}

From the perspective of \emph{generalization analysis}, the existing representative works on mmWave-based HPE and HMR still exhibit shortcomings, as shown in Table~\ref{tab:dataset_comparison}. Specifically, two significant limitations can be identified: (1) \textbf{limited diversity of measurement configurations}. Most existing datasets are collected under only one or a very limited number of measurement configurations, particularly lacking systematic variations in human position, human orientation, and subarray size. As a result, they are not suitable for generalization analysis with respect to measurement configuration shifts, since the necessary distribution coverage is absent at the data collection stage. (2) \textbf{restricted comparability for generalization analysis}. The existing mmWave-based HMR datasets do not fully disclose the raw I/Q data, and the precise radar spatial positions are not recorded either. This limits the community's ability to compare different algorithms, input representations, and modeling strategies in a broader and more fine-grained manner, and further constrains the study of generalization from raw measurements to mesh reconstruction. These limitations substantially hinder a comprehensive investigation of mmWave-based HMR under diverse sensing conditions.

To address these limitations, \textbf{DGHMesh}, a large-scale \textbf{d}ual-radar mmWave dataset and \textbf{g}eneralization-focused benchmark for \textbf{h}uman \textbf{mesh} reconstruction, is introduced. The dataset contains 8 actions and 360,000 frames, and supports modalities such as FMCW radar, SFCW radar, and synchronized RGB images. All annotations are generated through a high-precision 3D human reconstruction labeling system, ensuring geometric consistency and reliable supervision. To resolve restricted comparability for generalization analysis, precise radar spatial position information and synchronized raw I/Q data from both radar modalities are also recorded and released. In order to enhance the diversity of the limited measurement configurations, dynamic and static body conditions, multiple human positions, multiple human orientations, and different subarray sizes are all carefully considered during benchmark design, thereby offering the community a more flexible, accurate, and versatile benchmark for HMR research. Furthermore, to address the challenges associated with coupled regression of human mesh parameters in existing mmWave-based HMR methods and to validate the effectiveness of our proposed benchmark for generalization analysis, we present \textbf{mmPTM} (\textbf{mm}Wave \textbf{P}oint--\textbf{T}ube Causal \textbf{M}esh Former), a deep learning framework that jointly leverages FMCW point clouds and SFCW imaging tubes for mesh reconstruction. Extensive experiments demonstrate that the proposed method consistently achieves outstanding performance across multiple sub-benchmarks, confirming both the effectiveness of multi-radar fusion and the practical value of the proposed dataset and benchmark.

In summary, the main contributions of this study are as follows:

1) A high-precision 3D human mesh annotation system, along with a preprocessing pipeline for mmWave radar data, is developed to construct DGHMesh, a large-scale dual-radar mmWave dataset and generalization-focused benchmark for human mesh reconstruction.

2) In DGHMesh, synchronized raw I/Q data from dual mmWave radars, along with precisely calibrated radar spatial positions, are provided to overcome the restricted comparability for generalization analysis in existing HMR datasets, enabling more comprehensive and fair comparison of extensive algorithms.

3) A systematic HMR benchmark is established with diverse measurement configurations, including multiple human positions, human orientations, and subarray size settings, so as to mitigate the limited diversity of measurement configurations and facilitate generalization analysis under configuration-induced distribution shifts.

4) A query-based multi-radar fusion framework, mmPTM, is proposed to address the limitations of overemphasis on surface-level reconstruction accuracy and coupled parameter regression in existing methods for dynamic HMR.

\section{Related Works}

\subsection{mmWave-based Human Sensing Datasets and Benchmarks}
mmWave radar has led to two representative research directions in human sensing: HPE and HMR, as summarized in Table~\ref{tab:dataset_comparison}. Early studies mainly focused on HPE, i.e., recovering human keypoints or skeletal structures from radar signals. To promote algorithm development, several benchmarks and datasets have been introduced. Some works perform HPE with radar tensor representations, such as RF-Pose3D\cite{rfpose3d} and Hupr\cite{hupr}. In general, a radar tensor is a structured multi-dimensional representation derived from raw radar measurements through signal processing steps. Depending on the specific design, it can take the form of a range-doppler map, range-angle map, or a higher-dimensional radar cube, providing a compact representation of the spatial and motion characteristics of the target. Other works use radar point clouds for HPE, such as mmPose\cite{mmpose}, MARS\cite{mars}, mRI\cite{mri}, and mm-Fi\cite{mmfi}. Radar point clouds are sparse 3D point sets formed by estimating target scatterer locations from radar returns. Compared with tensor-based representations, point clouds preserve explicit spatial geometry and can further reduce the data volume. RT-Pose\cite{rtpose} was the first to provide raw I/Q data for HPE, enabling the simultaneous generation of radar tensor, point cloud, and other modalities. MM-DCDR\cite{mmdcdr} further introduced two mmWave radar devices for HPE and provided radar tensor and point cloud representations respectively, but did not conduct further research on multi-radar modal fusion. Nevertheless, existing HPE benchmarks and systems still rely on visual sensors or vision-assisted pipelines to generate annotations, and the resulting labels are often not sufficiently accurate, especially when radar and vision are not perfectly aligned or when the human body is heavily occluded or moving rapidly.

These limitations have motivated the community to move from HPE toward HMR. Unlike HPE, which focuses on sparse keypoints, HMR aims to reconstruct the full 3D human mesh and thus provides richer geometric information, including body shape, surface structure, and body-size variation, making it more suitable for fine-grained human perception tasks\cite{hmr1}. Research on mmWave-based HMR started later. In 2020, mmMesh\cite{mmmesh} first generated human mesh annotations using a motion capture system and applied a commercial mmWave radar to reconstruct dynamic human meshes, but the dataset was not publicly released. Later, mmBody\cite{mmbody} improved the task formulation in terms of motion diversity, mesh annotation generation, and the public availability of the dataset. More recently, M4Human\cite{m4human} provided both radar tensor and point cloud modalities and increased the dataset scale. 

Despite these advances, existing mmWave-based HMR studies and datasets still exhibit limited coverage of sensing conditions and incomplete exposure of raw measurement information. In particular, most datasets are collected under only a narrow range of human positions, human orientations, and array configurations, which makes it difficult to assess how accuracy of HMR changes as the sensing setup varies. This hinders a more systematic study of mmWave-based HMR under diverse sensing conditions.

\subsection{mmWave-based Human Mesh Reconstruction Methods}

HMR was originally developed in the computer vision community, where the input modality is typically one RGB image or short image sequence\cite{khan2022three}. Given the strong semantic content and dense appearance cues in images, early HMR methods primarily relied on CNN-based feature extraction and parametric human body models to infer 3D pose and shape from 2D observations\cite{smpl,cv1,cv2}. However, mmWave-based HMR is more challenging than image-based HMR due to the sparse, noisy, and modality-specific nature of radar measurements, as well as the absence of rich texture and appearance cues.

The first mmWave-based HMR study, mmMesh\cite{mmmesh}, explored the use of mmWave radar point clouds as input and proposed an anchor-clustering strategy together with an LSTM network to perform continuous dynamic human mesh reconstruction. This work demonstrated the feasibility of regressing body mesh from mmWave observations. Subsequently, M4esh\cite{m4esh} introduced a more structured representation pipeline by incorporating coarse-to-fine skeletal representations, point cloud filtering, and graph convolutional networks. More recently, P4Transformer\cite{mmbody} brought attention mechanisms and temporal step encoding into the HMR framework, further improving reconstruction performance by modeling both spatial dependencies and temporal dynamics. In addition, mmGPE\cite{mmgpe} investigated the use of virtual synthetic radar data for HMR, suggesting that simulation-based data augmentation may be beneficial for mmWave-driven mesh reconstruction.

Despite these advances, existing mmWave-based HMR methods are still dominated by single-modality or single-representation designs, such as radar point cloud or radar tensor. Most of them focus on directly regressing mesh-related parameters from a unified feature stream, without explicitly introducing a stronger structural prior for disentangling pose, shape, and translation. This leaves an important gap in the literature, especially for building HMR models that are both geometrically accurate and robust to sensing variations.

\section{mmWave-based 3D human mesh annotation system}

\begin{figure*}[t]

  \centering
  
  \includegraphics[width=0.8\textwidth]{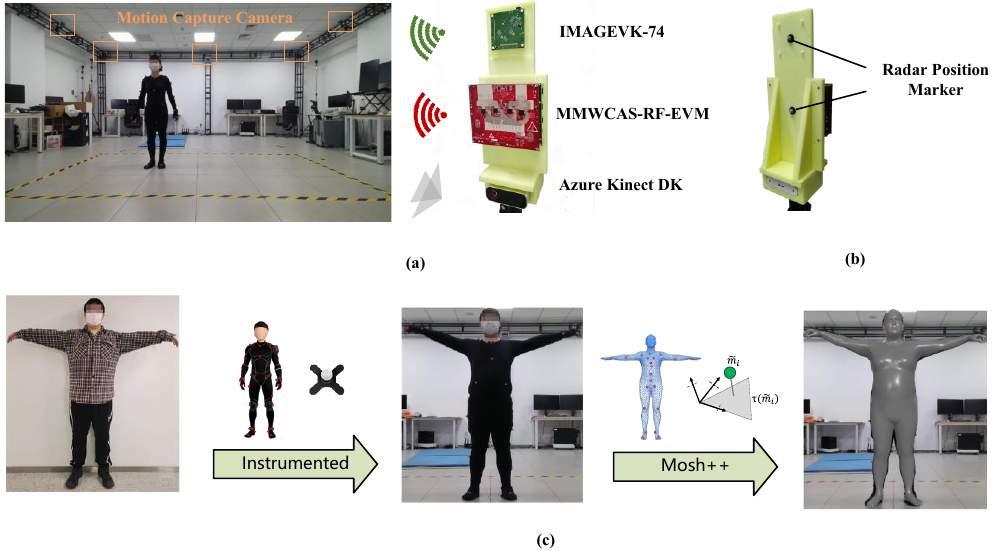}
  
  \caption{System overview. (a) Data acquisition hardware and experimental site; (b) The rear side of the acquisition fixture with reflective markers attached; (c) Generation pipeline of the human mesh ground truth.}

  \label{fig:system}
  
\end{figure*}

In this study, a multi-modal mmWave-based 3D human mesh annotation system is proposed, as illustrated in Fig.~\ref{fig:system}. During the experiments conducted with this system, subjects wear motion capture suits with markers attached according to predefined protocol, while FMCW radar data, SFCW radar data, RGB images and motion capture data are synchronously acquired. In addition, the precise positions of the radars are recorded. After data acquisition, a parameterized human body model fitting method based on motion capture markers is further employed to generate the human mesh as ground truth.

\subsection{Hardware System}

For the mmWave sensing hardware, the proposed system employs two mmWave radars with different modulation schemes, as shown in Fig.~\ref{fig:system}(a). The first is Texas Instruments' MMWCAS-RF-EVM\cite{ti}, an FMCW radar based on the AWR2243 cascade architecture, which supports up to 12 Tx and 16 Rx antenna elements, operates over the 76–81 GHz band, and provides approximately 1.4° azimuth resolution and 18° elevation resolution. The second is Mini-Circuits’ IMAGEVK-74\cite{minicircuit}, an SFCW radar with 20 Tx and 20 Rx on-board antennas, covering the 62–69 GHz band and designed for high-resolution 4D imaging. For camera capture, an Azure Kinect DK\cite{kinect} is used to collect synchronized RGB images, and its 12-MP RGB camera is employed in this study. For the MoCap system, a 16-camera OptiTrack\cite{mocap} setup with high-frame-rate infrared cameras is used. These infrared cameras provide sub-millimeter positional accuracy of approximately ±0.2 mm. Human pose supervision is obtained by tracking the retro-reflective markers attached to the human body, whose reflections are captured by the infrared cameras and reconstructed into precise pose information.

\subsection{Radar Position Marking}

During the experiments, a carefully designed 3D-printed fixture is used to rigidly mount the Azure Kinect and the two mmWave radars onto a tripod head, as shown in Fig.~\ref{fig:system}(b). To obtain the precise spatial positions of the radars during data collection, retro-reflective markers are also attached to the rear side of both radars, and their marker information is recorded synchronously throughout the data acquisition process. Based on the known geometry of the hardware and the dimensions of the fixture, the exact positions of all transmit and receive antennas on each radar can then be computed. In addition, to ensure that the tripod head remained level, a laser pointer and a spirit level are used for calibration before each experiment.

\subsection{Full Body Annotation}

To obtain the ground-truth human mesh, as shown in Fig.~\ref{fig:system}(c), all subjects first wear tight-fitting motion capture suits, including head caps and shoe covers, to ensure consistent marker attachment. Reflective markers are then placed on 39 body locations according to the standard Plug-in Gait marker placement protocol used in Vicon systems\cite{vicon}. In terms of details, cross-based reflective markers are directly attached to the suits, and additional hook-and-loop fasteners are used at loose-fitting regions to secure the markers and minimize displacement caused by garment motion during dynamic activities. After data collection, a neutral SMPL-X\cite{smplx} template is used as the reference model, and MoSh++\cite{mosh} is employed to fit the parametric body representation. The resulting triangular mesh and joint information are then taken as the ground truth. For the hand pose and facial expression parameters, the default template values are adopted and kept fixed throughout the fitting process. As shown in Fig.~\ref{fig:system}(c), the ground-truth mesh is well aligned with the RGB images.

\subsection{Modal Synchronization}

During data acquisition, the radars, RGB, and MoCap streams are synchronized and recorded on the same computer to ensure temporal consistency across modalities. The timestamp of each frame of raw radar data, each RGB image, and each MoCap recording is logged, enabling millisecond-level temporal alignment. All modalities are further temporally aligned to a unified frame rate of 10 fps. In this way, the collected multi-modal data provide reliable supervision for annotation generation, benchmark construction, and cross-modal fusion.

\section{Methods}

\begin{figure*}[t]

  \centering
  
  \includegraphics[width=\textwidth]{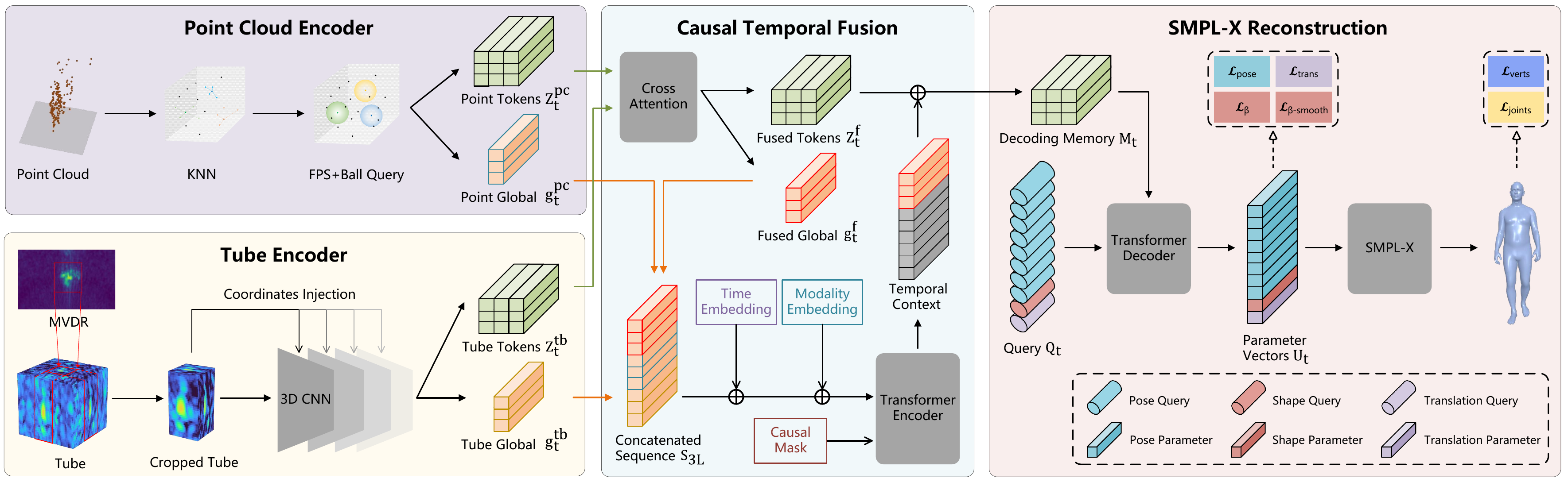}
  
 \caption{Overview of the proposed mmPTM. The FMCW radar point cloud is encoded by the point cloud encoder into point tokens $Z_t^{\mathrm{pc}}$ and a global token $g_t^{\mathrm{pc}}$, while the cropped SFCW radar tube is processed by the tube encoder into tube tokens $Z_t^{\mathrm{tb}}$ and a global token $g_t^{\mathrm{tb}}$. In causal temporal fusion, cross-attention yields fused tokens $Z_t^{\mathrm{f}}$ and a fused global token $g_t^{\mathrm{f}}$, from which the decoding memory $M_t$ is obtained by injecting temporal context extracted by a causal Transformer encoder. Finally, $M_t$ and the query set $Q_t$ are fed into the SMPL-X decoder to regress pose, shape, and translation parameters, which are used to reconstruct the human mesh vertices and joints.}

  \label{fig:mmptm}
  
\end{figure*}

\subsection{Overall}

In this section, we present the overall framework for multimodal mmWave-based 3D human mesh reconstruction. The proposed framework consists of two components: a preprocessing pipeline for mmWave radar data and a neural network for HMR named mmPTM. Given a sequence segment of length $L$, the raw FMCW measurements are converted into the point cloud sequence $\mathbf{X}_{\mathrm{pc}}\in\mathbb{R}^{L\times N_{\mathrm{pc}}\times 6}$, where $N_{\mathrm{pc}}$ denotes the number of point clouds per frame, and the 6-dimensional feature vector includes not only the 3D coordinates $(x,y,z)$ but also the radial velocity, reflected energy, and range value. In parallel, the raw SFCW measurements are transformed into the cropped 4D imaging tube $\mathbf{X}_{\mathrm{tb}}\in\mathbb{R}^{L\times N_x'\times N_y\times N_z'}$, where $N_x'$, $N_y$, and $N_z'$ denote the three spatial dimensions of the cropped tube. These two modality-specific representations are then jointly fed into mmPTM, which regresses SMPL-X parameters and reconstructs dynamic human meshes $\hat{\mathbf{V}}_t\in\mathbb{R}^{10475\times 3}$ and joints $\hat{\mathbf{J}}_t\in\mathbb{R}^{22\times 3}$.

\textbf{Preprocessing Pipeline for mmWave Radar Data.} 
To obtain consistent and informative radar representations for the subsequent mmPTM network, we design a preprocessing pipeline for the two mmWave modalities. For the FMCW radar, a candidate-peak-based generation procedure is used to construct the per-frame point cloud representation, thereby improving the reliability of point selection. For the SFCW radar, a back-projection-based 3D imaging procedure is first applied to obtain volumetric responses, and a sequence-level ROI together with per-frame tracking-box masking is then used to crop the imaging tube, which reduces background redundancy.

\textbf{mmPTM.} 
To address the challenges of overemphasis on surface-level reconstruction accuracy and coupled parameter regression in existing mmWave-based HMR methods, we propose mmPTM. As shown in Fig.~\ref{fig:mmptm}, the network is composed of four parts: point cloud encoder, tube encoder, causal temporal fusion, and SMPL-X reconstruction. 
The point cloud branch captures explicit geometric and kinematic cues from sparse radar returns, while the tube branch models dense volumetric energy patterns. The fusion and temporal aggregation modules then integrate the complementary information across modalities and over time, enabling causal sequence modeling for dynamic mesh reconstruction. Finally, the predicted SMPL-X parameters from a query-based decoder are fed into a frozen SMPL-X layer to generate the 3D meshes and joints.
Compared with existing LSTM-based radar HMR methods such as mmMesh~\cite{mmmesh} and M4esh~\cite{m4esh}, which mainly emphasize vertex-level and joint-level fitting, mmPTM places additional emphasis on articulated pose accuracy and temporal smoothness of human shape within each sequence window. Compared with Transformer-based regression methods such as P4Transformer~\cite{mmbody}, which directly predict all SMPL-X parameters in a single coupled regression space, mmPTM introduces a more structured inductive bias by explicitly disentangling pose, shape, and translation. Specifically, these factors are modeled by separate query groups and decoded from a shared memory feature, which reduces parameter coupling and mitigates gradient interference among different reconstruction targets.

The remainder of this section first details the preprocessing pipeline for the two radar modalities, and then presents the architecture and implementation details of mmPTM.

\subsection{Preprocessing Pipeline for mmWave Radar Data}
\textbf{Point Cloud Generation}. 
We generate point clouds from cascaded TDM--MIMO mmWave radar complex baseband data on a per-frame basis. For each frame, the measurements are arranged as a 4-D radar cube $x[t,r,\ell,n]$, where $t=1,\dots,T_x$ and $r=1,\dots,R_x$ index the transmit and receive antennas, $\ell$ indexes chirp loops within the frame, and $n$ indexes ADC samples. A range FFT along the fast-time axis yields the range spectrum $X_R[t,r,\ell,k]$. To suppress static background reflections, clutter removal is applied along the slow-time axis. A Doppler FFT is then performed along $\ell$ to obtain $X_D[t,r,m,k]$, where $m$ denotes the Doppler bin. Candidate target cells are localized via a Doppler--range (RD) energy map by summing over antenna channels:
\begin{equation}
E(m,k) \;=\; \sum_{t=1}^{T_x}\sum_{r=1}^{R_x}\left|X_D[t,r,m,k]\right|^2,
\end{equation}
and the top-$K$ RD peaks are selected as initial detections. Conventional FFT-based pipelines typically output only one 3D point per RD cell by taking a single strongest AoA peak~\cite{mmmesh}, which captures dominant human dynamics but may discard weaker yet informative body details at similar ranges. To mitigate this ''one-cell--one-point'' limitation, we adopt a candidate-peak mechanism that extracts multiple AoA local maxima per RD cell and consolidates them into a point set via unified scoring and gating. Specifically, for each selected RD cell, we detect multiple local peaks on the corresponding 2-D AoA spectrum and treat them as candidate spatial points, rather than keeping only the global maximum. These candidates are then ranked using an energy-related score that jointly reflects the RD cell strength and the AoA peak magnitude, while additional gating is applied to suppress implausible or redundant candidates. 

Finally, we set the number of point cloud points per frame within the effective space to $N_{\mathrm{pc}}=256$, producing $P\in\mathbb{R}^{256\times 6}$ with feature order $(x,y,z,v_r,E_{\mathrm{dB}},R)$, where $(x,y,z)$ denote the 3D spatial coordinates of each point cloud, $v_r$ represents the radial velocity, $E_{\mathrm{dB}}$ denotes the reflected signal energy in dB, and $R$ indicates the range value.

\textbf{4D Imaging Tube}. 
The SFCW radar 4D imaging tube is generated by a 3D imaging algorithm based on coherent back-projection. Let the SFCW radar have $T_{x,s}$ transmit antennas and $R_{x,s}$ receive antennas, yielding $M_s=T_{x,s}R_{x,s}$ virtual Tx--Rx channels. For each frame $t$, complex frequency responses are measured on a set of uniformly spaced tones $\{f_q\}_{q=0}^{Q-1}$, denoted by $y_t[m,q]$ for channel $m$ and tone $q$. We first apply a frequency window and perform an IFFT along $q$ to obtain a per-channel range profile $g_t[m,n]$. With tone spacing $\Delta f$, the range sampling interval is
\begin{equation}
\Delta \rho \;=\; \frac{c}{2N_{\mathrm{fft}}\Delta f},
\end{equation}
where $c$ is the speed of light and $N_{\mathrm{fft}}$ is the IFFT length.
Then, for each voxel location $\mathbf{r}=(x,y,z)$ on the 3D scan grid, we coherently back-project all channels. For the $i$-th transmitter and $j$-th receiver, the bistatic path length is $d_{i,j}$ and the corresponding equivalent range is $\rho(\mathbf{r})=d_{i,j}(\mathbf{r})/2$. We sample the range profile at $n\approx \rho(\mathbf{r})/\Delta\rho$ and apply a phase compensation term with $k_0=2\pi f_0/c$, where $f_0$ is the start tone frequency. The voxel value is computed by coherent summation:
\begin{equation}
V_t(\mathbf{r}) \;=\; \sum_{m=1}^{M_s} g_t\!\left[m,\frac{\rho_m(\mathbf{r})}{\Delta\rho}\right]\,
\exp\!\big(jk_0 d_m(\mathbf{r})\big),
\end{equation}
yielding a 3D volume $V_t(x,y,z)$ per frame. Stacking all frames forms the initial 4D imaging tube with tensor shape $(T,N_x,N_y,N_z)$, where $N_x$, $N_y$, and $N_z$ are the scan-grid sizes.

Since human perception in a large 3D space contains substantial irrelevant background voxels, feeding the full tube into the neural network is inefficient. Therefore, we use the cascaded TDM--MIMO MVDR heatmap\cite{mvdr} to generate per-frame human tracking boxes and crop the SFCW tube accordingly. Specifically, we estimate a per-frame human tracking bounding box (bbox) on the $x$--$z$ plane from the MVDR heatmap, and use it to crop the SFCW 4D imaging tube along $x$ and $z$ while keeping the full $y$ axis. The tracking algorithm takes the per-frame MVDR heatmap as input, converts it to linear power, and suppresses static background via a column-wise baseline removal. It then aggregates responses along $z$ to form a 1D energy profile over $x$, on which an OS-CFAR\cite{oscfar} detector is applied to obtain candidate peaks. Around the selected peak, a local $x$--$z$ neighborhood is analyzed by extracting the connected high-response region and estimating the human center $(x_t,z_t)$ using weighted statistics, followed by temporal smoothing to reduce jitter. The per-frame bbox is defined by a fixed physical box size $B$ centered at $(x_t,z_t)$. This bbox is then used to guide the subsequent spatial cropping of the SFCW tube.

Finally, using a sequence-level union ROI followed by per-frame bbox masking, we obtain the cropped 4D imaging tube with tensor shape $(T,N_x',N_y,N_z')$, where $N_x'$ and $N_z'$ are significantly smaller than the original $N_x$ and $N_z$.

\subsection{mmPTM}

The inputs to mmPTM follow the previously defined notation: each frame produces $P\in\mathbb{R}^{256\times 6}$ with feature order $(x,y,z,v_r,E_{\mathrm{dB}},R)$. We use a sliding window of length $L$ as the network input, denoted by $X_{\mathrm{pc}}\in\mathbb{R}^{L\times N_{\mathrm{pc}}\times 6}$.
For the 4D imaging tube, the sliding window of cropped tube is organized as $X_{\mathrm{tb}}\in\mathbb{R}^{L\times N_x'\times N_y\times N_z'}$, together with metadata: crop center $\mathbf{c}_t=(x_{c,t},z_{c,t})\in\mathbb{R}^{2}$ and physical axis samples $x_{\mathrm{scan}}\in\mathbb{R}^{L\times N_x'}$, $y_{\mathrm{scan}}\in\mathbb{R}^{N_y}$, $z_{\mathrm{scan}}\in\mathbb{R}^{L\times N_z'}$.

\subsubsection{Point Cloud Encoder}
The point cloud encoder maps the sparse and unordered FMCW point cloud window
$X_{\mathrm{pc}}$
into per-frame point tokens and a global token.
The key design motivation is that mmWave point cloud returns are noisy, incomplete, and non-uniformly distributed; thus, neighborhood aggregation should not rely solely on feature similarity, but should be explicitly informed by relative geometry and radar-attribute changes to stabilize local body structure and motion cues.

For each frame $t$, let $\mathbf{x}_i\in\mathbb{R}^{3}$ denote the 3D coordinate of point $i$, and
$\mathbf{e}_i\in\mathbb{R}^{3}$ denote its non-coordinate radar attributes (e.g., $v_r,E_{\mathrm{dB}},R$).
And let $\mathbf{h}_i\in\mathbb{R}^{d}$ denote the learnable feature embedding associated with point $i$ at the input of the local aggregation.
We build a K-Nearest Neighbor (KNN) graph in the coordinate space and perform relation-aware local attention over the neighborhood
$\mathcal{N}_k(i)$ of each point $i$.

For each edge $(i,j)$ with $j\in\mathcal{N}_k(i)$, we construct a relation descriptor that explicitly encodes geometry and radar-attribute variations:
\begin{equation}
\mathbf{r}_{ij}=\Big[\;(\mathbf{x}_j-\mathbf{x}_i),\ \|\mathbf{x}_j-\mathbf{x}_i\|_2,\ (\mathbf{e}_j-\mathbf{e}_i)\;\Big].
\end{equation}
This relation is mapped by an MLP $\psi(\cdot)$ and injected into the neighbor key/value representations,
\begin{equation}
\mathbf{k}_{ij}=\mathbf{W}_k\mathbf{h}_j+\psi(\mathbf{r}_{ij}),\quad
\mathbf{v}_{ij}=\mathbf{W}_v\mathbf{h}_j+\psi(\mathbf{r}_{ij}),
\end{equation}
while the query is computed from the center feature $\mathbf{q}_i=\mathbf{W}_q\mathbf{h}_i$.
A relation-aware local attention aggregates information over $\mathcal{N}_k(i)$:
\begin{equation}
\tilde{\mathbf{h}}_i=\mathbf{h}_i+\sum_{j\in\mathcal{N}_k(i)}
\mathrm{softmax}_j\!\left(\frac{\mathbf{q}_i^\top \mathbf{k}_{ij}}{\sqrt{d}}\right)\mathbf{v}_{ij}.
\end{equation}
After local attention , we obtain the intermediate point set represented by coordinates $\tilde{\mathbf{P}}_t\in\mathbb{R}^{N_{\mathrm{pc}}\times 3}$ and locally aggregated features $\tilde{\mathbf{H}}_t\in\mathbb{R}^{N_{\mathrm{pc}}\times d}$

To enlarge the receptive field and reduce the point set for downstream modules, we apply two-stage Set Abstraction (SA):
Farthest Point sampling (FPS) selects center points and ball query gathers radius neighborhoods at each level.
After two SA levels, we obtain per-frame point tokens
$Z_t^{\mathrm{pc}}\in\mathbb{R}^{N_{\mathrm{pc}}'\times d'}$.
Finally, an attention pooling layer produces a compact frame-level global token
$\mathbf{g}_t^{\mathrm{pc}}\in\mathbb{R}^{d'}$ for subsequent cross-modal fusion and causal temporal aggregation.

\subsubsection{Tube Encoder}
ROI cropping reduces volumetric computation but weakens absolute location cues in the global coordinate system,
whereas directly feeding per-voxel absolute coordinates $(x_{\mathrm{scan}},y_{\mathrm{scan}},z_{\mathrm{scan}})$ is highly redundant.
We therefore restore the necessary position information with a compact \emph{coordinate injection + global conditioning} design.

We use a 4-stage strided 3D CNN to encode the cropped tube volume.
At each stage, we inject a lightweight physical-coordinate bias by resampling the 1D axes
$x_{\mathrm{scan}},y_{\mathrm{scan}},z_{\mathrm{scan}}$ to the current feature resolution and mapping them with learnable linear layers
into per-channel bias terms that are broadcast and added to the feature map.
This provides absolute-position cues without concatenating a full coordinate grid to the input.

We further apply FiLM to re-calibrate channel responses per frame.
We form a conditioning vector $\mathbf{s}_t\in\mathbb{R}^{5}$ by concatenating the crop center
$\mathbf{c}_t=(x_{c,t},z_{c,t})\in\mathbb{R}^{2}$ and fixed reference scales, and predict
$\gamma(\mathbf{s}_t),\beta(\mathbf{s}_t)\in\mathbb{R}^{C}$ with linear layers:
\begin{equation}
\mathrm{FiLM}(F;\mathbf{s}_t)=F\odot(1+\gamma(\mathbf{s}_t))+\beta(\mathbf{s}_t),
\end{equation}
where $\odot$ denotes channel-wise multiplication.
Finally, we flatten the deepest feature volume into a voxel-token sequence and project it to dimension $d'$,
yielding tube tokens $Z_t^{\mathrm{tb}}\in\mathbb{R}^{N_v\times d'}$ (where $N_v$ denotes the \emph{voxel-token resolution} of the deepest tube feature volume),
and obtain a global token $\mathbf{g}_t^{\mathrm{tb}}\in\mathbb{R}^{d'}$ via attention pooling.

\subsubsection{Causal Temporal Fusion}
We align the two modalities by multi-head cross-attention from point tokens to tube tokens.
Given point tokens $Z_t^{\mathrm{pc}}\in\mathbb{R}^{N_{\mathrm{pc}}'\times d'}$ (queries) and tube tokens
$Z_t^{\mathrm{tb}}\in\mathbb{R}^{N_v\times d'}$ (keys/values), a cross-attention block produces a tube-aware refinement
$\Delta Z_t\in\mathbb{R}^{N_{\mathrm{pc}}'\times d'}$.
Since tube measurements may contain ghost structures, we predict a per-frame reliability
gate from the tube global descriptor $\mathbf{g}_t^{\mathrm{tb}}$ to suppress negative transfer:
\begin{equation}
q_t=\sigma(\mathrm{MLP}(\mathbf{g}_t^{\mathrm{tb}})),\qquad
Z_t^{\mathrm{f}}=Z_t^{\mathrm{pc}}+q_t\,\Delta Z_t,
\end{equation}
where $q_t\in(0,1)$ is a scalar applied to all $N_{\mathrm{pc}}'$ tokens of frame $t$.
We then obtain a fused global token $\mathbf{g}_t^{\mathrm{f}}\in\mathbb{R}^{d'}$ by attention pooling over
$Z_t^{\mathrm{f}}$.

To enforce temporal consistency without future leakage, we interleave three per-frame descriptors into a length-$3L$
sequence
\begin{equation}
S_{3L}=[\mathbf{g}_1^{\mathrm{pc}},\mathbf{g}_1^{\mathrm{tb}},\mathbf{g}_1^{\mathrm{f}},\ldots,
\mathbf{g}_L^{\mathrm{pc}},\mathbf{g}_L^{\mathrm{tb}},\mathbf{g}_L^{\mathrm{f}}]\in\mathbb{R}^{3L\times d'}.
\end{equation}
We add a time embedding and a modality-type embedding to each element of $S_{3L}$ and feed the resulting sequence to
a Transformer encoder with an upper-triangular causal mask.
We then extract the outputs at the fused positions to obtain per-frame temporal context
$\mathbf{u}_t\in\mathbb{R}^{d'}$. 
Finally, we inject $\mathbf{u}_t$ to $Z_t^{\mathrm{f}}$, and form the decoding memory $M_t$.

\subsubsection{SMPL-X Reconstruction}
We map the per-frame memory $M_t\in\mathbb{R}^{N_{\mathrm{pc}}'\times d'}$ to interpretable SMPL-X parameters using a
query-based decoder, followed by a frozen and differentiable SMPL-X body layer. We introduce
learnable semantic queries: $J$ pose queries $\{\mathbf{q}_j\}_{j=1}^{J}$, a translation query
$\mathbf{q}_{\mathrm{trans}}$, and a shape query $\mathbf{q}_{\beta}$. The initial query set is
\begin{equation}
Q_t=[\mathbf{q}_1,\ldots,\mathbf{q}_J,\mathbf{q}_{\mathrm{trans}},\mathbf{q}_{\beta}]
\in\mathbb{R}^{(J+2)\times d'},
\end{equation}
where each $\mathbf{q}\in\mathbb{R}^{d'}$ is a learnable vector. Using semantic queries encourages a structured readout
from $M_t$ (joint-wise pose, global translation, and shape) and reduces parameter coupling compared to directly
regressing all parameters with single MLP. We update $Q_t$ with a query-to-memory attention decoder
to obtain $U_t$.

Linear heads regress joint rotations in a 6D representation $\hat{\mathbf{p}}_{t,j}\in\mathbb{R}^{6}$, translation
$\hat{\mathbf{t}}_t\in\mathbb{R}^{3}$, and shape $\hat{\boldsymbol{\beta}}_t\in\mathbb{R}^{D_{\beta}}$.
Since common 3D rotation parameterizations are discontinuous for neural networks\cite{zhou2019continuity}, we adopt the continuous 6D rotation
representation and convert it to rotation matrices via
\begin{equation}
\hat{\mathbf{R}}_{t,j}=\Pi_{\mathrm{6D}\rightarrow SO(3)}(\hat{\mathbf{p}}_{t,j})\in SO(3),
\end{equation}
where $\Pi_{\mathrm{6D}\rightarrow SO(3)}(\cdot)$ denotes the continuous mapping from 6D to a $3\times 3$ rotation
matrix. The SMPL-X layer outputs mesh vertices and joints:
\begin{equation}
(\hat{V}_t,\hat{J}_t)=\mathrm{SMPLX}(\hat{\boldsymbol{\beta}}_t,\hat{\mathbf{R}}_{t,1:J},
\hat{\mathbf{t}}_t),
\end{equation}
where $\hat{V}_t\in\mathbb{R}^{10475\times 3}$ for the default SMPL-X topology, and we use the first $J=22$ joints from
$\hat{J}_t$ for supervision/evaluation.

\begin{figure*}[t]

  \centering
  
  \includegraphics[width=0.9\textwidth]{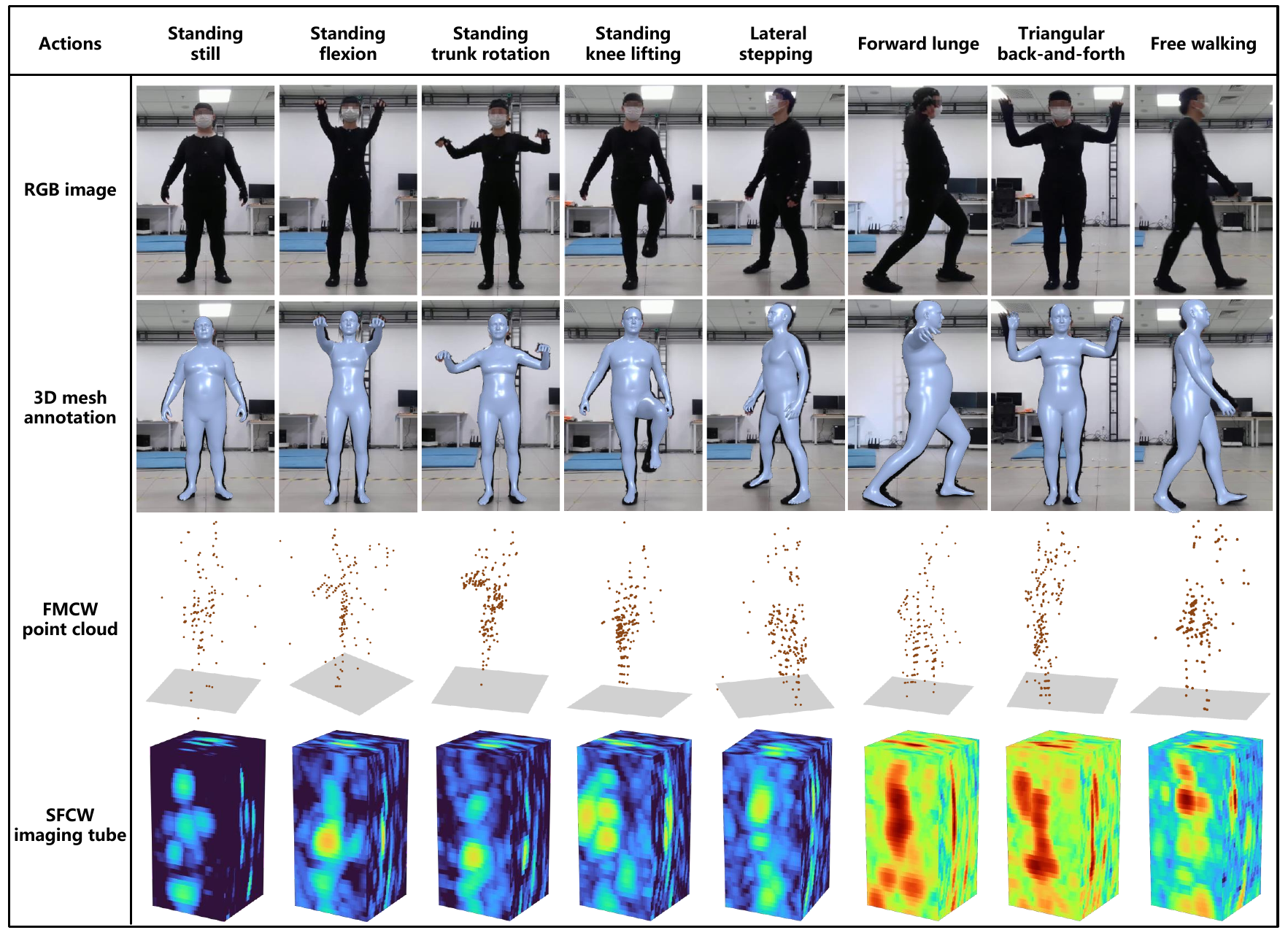}
  
 \caption{Visualizations of synchronized multimodal data in DGHMesh. For each action, the RGB image, 3D mesh annotation, FMCW point cloud, and SFCW imaging tube are presented in a time-aligned manner.}

  \label{fig:dataset}
  
\end{figure*}

\subsection{Model Loss}
Our training objective is a weighted sum of losses:
\begin{equation}
\begin{aligned}
\mathcal{L}&=\\
& \lambda_{\mathrm{verts}}\underbrace{\frac{1}{|\mathcal{V}|}\sum_{i\in\mathcal{V}}\left\lVert \hat{V}_{L,i}-V_{L,i}\right\rVert_2}_{\mathcal{L}_{\mathrm{verts}}}
+\lambda_{\mathrm{joints}}\underbrace{\frac{1}{J}\sum_{j=0}^{J-1}\left\lVert \hat{J}_{L,j}-J_{L,j}\right\rVert_2}_{\mathcal{L}_{\mathrm{joints}}} \\
&+\lambda_{\mathrm{pose}}\underbrace{\frac{1}{J}\sum_{j=0}^{J-1}\theta\!\left(\hat{\mathbf{R}}_{L,j}^{\top}\mathbf{R}_{L,j}\right)}_{\mathcal{L}_{\mathrm{pose}}}
+\lambda_{\beta}\underbrace{\mathrm{SmoothL1}\!\left(\hat{\boldsymbol{\beta}}_{L},\boldsymbol{\beta}_{L}\right)}_{\mathcal{L}_{\beta}} \\
&+\lambda_{\beta\text{-smooth}}\underbrace{\frac{1}{L-1}\sum_{t=2}^{L}\left\lVert \hat{\boldsymbol{\beta}}_{t}-\hat{\boldsymbol{\beta}}_{t-1}\right\rVert_2^{2}}_{\mathcal{L}_{\beta\text{-smooth}}}
+\lambda_{\mathrm{trans}}\underbrace{\left\lVert \hat{\mathbf{t}}_{L}-\mathbf{t}_{L}\right\rVert_2}_{\mathcal{L}_{\mathrm{trans}}}.
\end{aligned}
\end{equation}

Here $L$ is the window length, and supervision is mainly applied to the last frame $t=L$. The coefficients $\lambda_{\mathrm{verts}},\lambda_{\mathrm{joints}},\lambda_{\mathrm{pose}},\lambda_{\beta},
\lambda_{\beta\text{-smooth}},\lambda_{\mathrm{trans}}$ are scalar hyperparameters that weight different loss terms.
$\hat{V}_{L}$ and $\hat{J}_{L}$ denote the predicted SMPL-X mesh vertices and joints at $t=L$, with ground truth
$V_{L}$ and $J_{L}$.
$\{\hat{\mathbf{R}}_{L,j}\}$ and $\{\mathbf{R}_{L,j}\}$ denote the predicted and ground-truth joint rotation matrices,
and $\theta(\cdot)$ denotes the geodesic rotation error on $SO(3)$.
$\hat{\boldsymbol{\beta}}_{L}$ denotes the predicted shape at $t=L$, while
$\hat{\boldsymbol{\beta}}_{t}$ denotes the predicted shape sequence within the window for the smoothness term
$\mathcal{L}_{\beta\text{-smooth}}$.
Finally, $\hat{\mathbf{t}}_{L}$ and $\mathbf{t}_{L}$ denote the predicted and ground-truth translation at $t=L$.

\section{Benchmark and Evaluation}

\subsection{Benchmark}

\subsubsection{Dataset Descriptions}\hfill\break
\indent\textbf{Subjects.}
A total of 15 volunteers with diverse body shapes were recruited for dataset collection, including 8 males and 7 females. Their heights ranged from 155 cm to 187 cm, and their weights ranged from 46 kg to 105 kg. Before the experiments, all participants signed informed consent forms, were fully informed of the experimental procedure, and were notified that they could withdraw from the study at any time. In addition, all participants were required to wear masks during data collection to ensure anonymization and protect their privacy.

\begin{figure}[t]

  \centering
  
  \includegraphics[width=0.8\columnwidth]{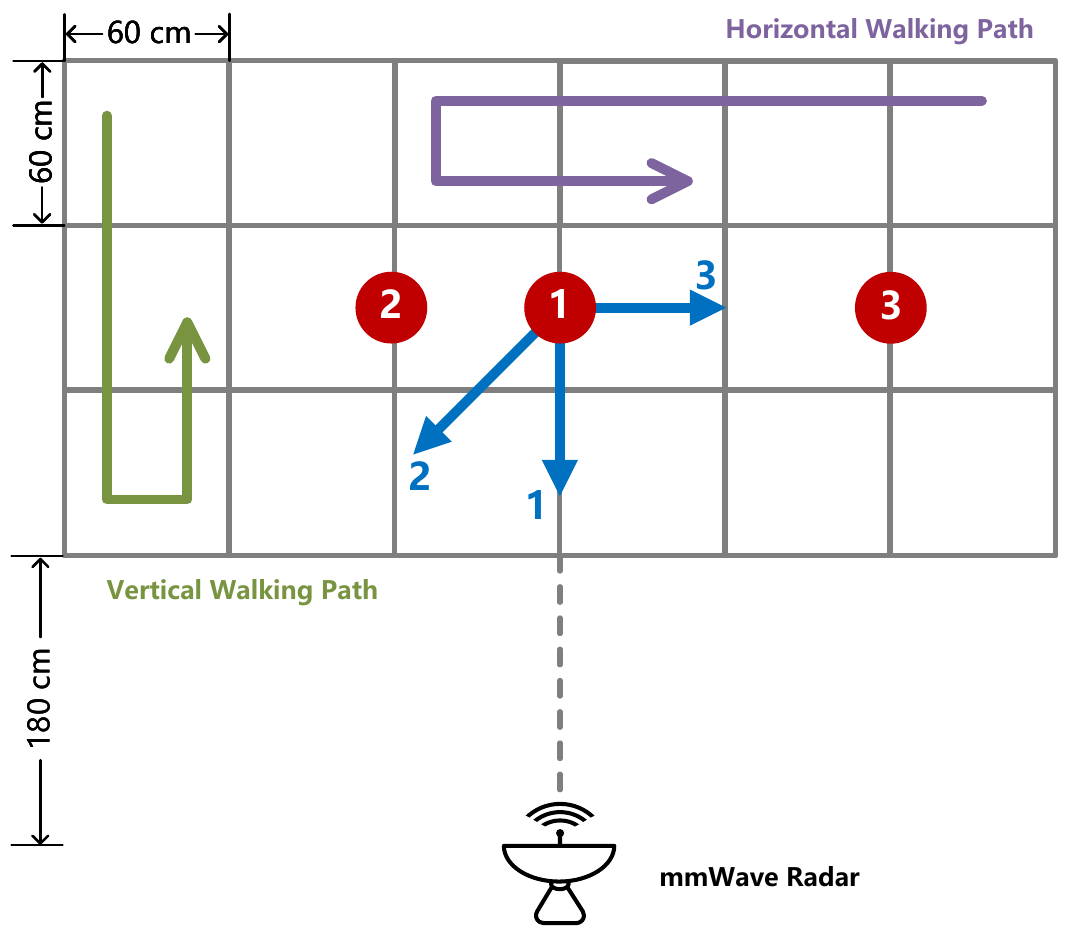}
  
  \caption{Action Sequences. The gray grids denote the activity area of the subject. The red-marked locations indicate the subject’s positions in the three sequences of Action Category I, with the boresight of the mmWave radar aligned toward Position 1. The blue-marked arrows represent the subject’s orientations in the three sequences of Action Category II. The green and purple trajectories illustrate the vertical and horizontal grid walking paths, respectively, in Action Category III.}

  \label{fig:actions}
  
\end{figure}

\textbf{Action Sequences.}
To support sub-benchmark studies, each subject was asked to perform three categories of actions, covering a total of eight motion patterns: actions involving global position changes (Actions Category I), actions involving changes in human orientation (Actions Category II), and walking actions (Actions Category III). Specifically, the Actions Category I included four motions, namely standing still, standing flexion, standing trunk rotation, and standing knee lifting. The Actions Category II included three motions, namely lateral stepping, forward lunge, and triangular back-and-forth. In addition, Actions Category III includes one free walking action. Fig~\ref{fig:dataset} presents the synchronized RGB images, 3D mesh annotations, FMCW point clouds, and SFCW imaging tubes for all actions in DGHMesh.

For each action, three sequences were performed. As shown in Fig~\ref{fig:actions}, for the Actions Category I, the subject’s global position sequentially appeared at the three red-marked locations labeled 1, 2, and 3. For the Actions Category II, the subject’s body orientation sequentially followed the three blue-marked directions labeled 1, 2, and 3. For the Actions Category III, the subject sequentially walked along a vertical grid path, a horizontal grid path, and a random path. In total, each subject performed 8 actions and 24 sequences. Since each sequence lasted 100 s, the total duration of the entire dataset was approximately 10 h.

\textbf{Radar Configuration Parameters.}
For the MMWCAS-RF-EVM, the RF starting frequency was set to 77.0 GHz, with a frequency slope of \SI{60.012}{MHz/\micro\second} and 64 loops per frame. Under this configuration, the resulting range resolution was approximately 4.3 cm, and the velocity resolution was approximately 3.5 cm/s. For the IMAGEVK-74, the RF sweep started at 63.0 GHz and ended at 66.4 GHz, with a total of 128 frequency sampling points. Based on this configuration, the achieved range resolution was approximately 4.4 cm. This parameter configuration achieves a good balance between spatial resolution and temporal resolution, and is suitable for the human motion perception and mesh reconstruction tasks in this study.

\begin{table*}[t]
\centering
\caption{Summary of the proposed sub-benchmarks.}
\label{tab:benchmark_construction}
\begin{tblr}{
  cells = {valign=m},
  colspec = {
    Q[l,m]
    Q[l,m]
    Q[l,m]
    Q[l,m]
    Q[l,m,wd=0.3\textwidth]
  },
  row{1} = {font=\bfseries},
  hline{1,Z} = {-}{0.1em},
  hline{2} = {-}{0.03em},
}
Sub-benchmark               & Variation Factor               & Action       & Subject       & Test Purpose                                                                    \\
Human position variation   & Human global displacement                 & Actions Category I  & Within-subject   & Quantify how radar viewpoint shifts (induced by the changes in the relative position between the radar and the human body) affects human mesh reconstruction (HMR) robustness.                \\
Human orientation variation & Human facing direction~        & Actions Category II & Within-subject   & Evaluate how stable reconstruction is under body orientation changes. \\
Subarray size variation     & Array aperture                 & All actions         & Within-subject   & Study how array aperture impacts reconstruction accuracy and its sensitivity across different motion patterns.                   \\
Subject-specific            & Subject identity               & All actions         & Cross-subject & Measure how well methods generalize to unseen subjects (body shape and motion style) rather than overfitting identity cues.   \\
\end{tblr}
\end{table*}

\subsubsection{Benchmark Construction}\hfill\break
\indent The proposed benchmark is designed to systematically evaluate the generalization capability of mmWave-based HMR methods under different measurement configurations. Rather than focusing on a single sensing condition, the benchmark is organized around four complementary sub-benchmarks that jointly probe the major factors affecting mmWave-based human mesh reconstruction, including human position, human orientation, array resolution, and subject identity. As summarized in Table~\ref{tab:benchmark_construction}, each sub-benchmark is constructed to isolate one specific factor while keeping the remaining conditions as consistent as possible, so that its impact on reconstruction accuracy and robustness can be independently assessed.

\textbf{Human Position Variation Sub-benchmark.}
In practice, changing the relative position between the radar and human body alters the viewing geometry, line-of-sight coverage, and angular sampling of the human body, which in turn changes the structure and sparsity of the received radar measurements. This sub-benchmark therefore attempts to evaluate: \textbf{\textit{how robust are mmWave-based HMR methods to radar viewpoint shifts?}}

\textbf{Human Orientation Variation Sub-benchmark.}
Different body orientations produce different radar cross-section patterns, self-occlusion levels, and reflection distributions. This sub-benchmark therefore attempts to evaluate: \textbf{\textit{how stable are mmWave-based HMR methods when body orientation changes?}}

\begin{figure}[t]

  \centering
  
  \includegraphics[width=\columnwidth]{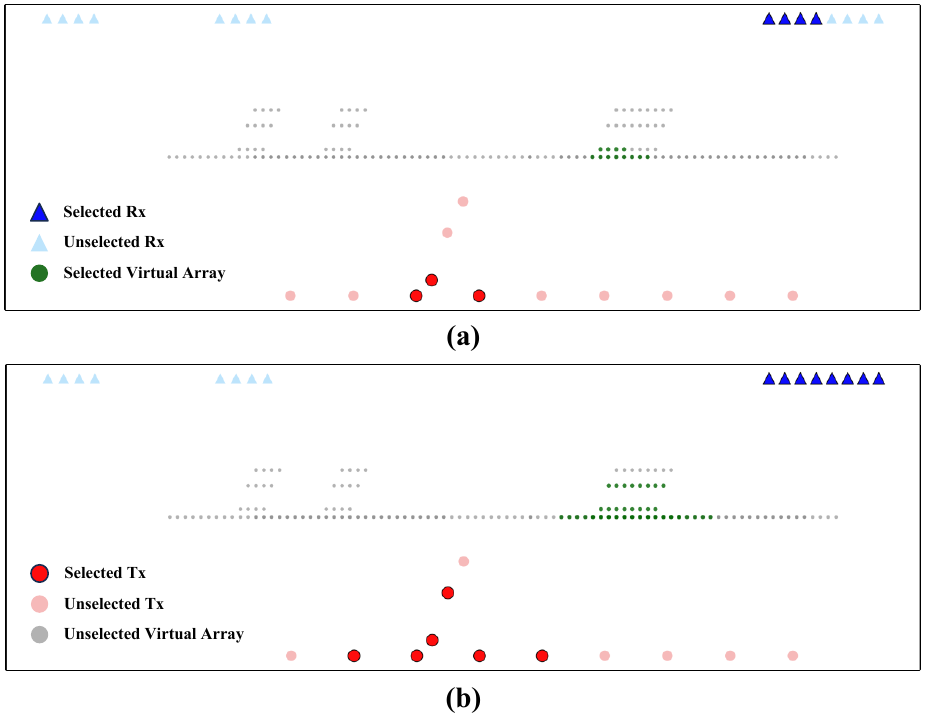}
  
  \caption{Subarray Configuration. Red circles denote the selected Tx, while light red circles indicate the unselected Tx. Blue triangles represent the selected Rx, and light blue triangles denote the unselected Rx. The green circles correspond to the virtual array elements formed by the selected Tx–Rx pairs, whereas the gray circles indicate the unselected virtual array elements. (a) Subarray with 3 Tx $\times$ 4 Rx. (b) Subarray with 6 Tx $\times$ 8 Rx.}

  \label{fig:subarray}
  
\end{figure}

\textbf{Subarray Size Variation Sub-benchmark.}
In mmWave sensing, the subarray size directly influences angular resolution and spatial discrimination capability. Beyond reconstruction accuracy, varying the subarray size also provides a practical means to examine the trade-off among performance, cost-efficiency, and portability. A smaller subarray is generally associated with reduced hardware cost, lower power consumption, and improved portability, but at the expense of degraded sensing resolution. In contrast, a larger array has the opposite effect. This sub-benchmark therefore attempts to evaluate: \textbf{\textit{how does array aperture impact mmWave-based HMR?}}

In this benchmark, two subarray configurations are considered for the FMCW radar, namely 3 transmit $\times$ 4 receive (3$\times$4) and 6 transmit $\times$ 8 receive (6$\times$8). The full FMCW radar array supports up to 12 transmit and 16 receive channels, and the selected subarrays are extracted from the complete array to simulate different sensing budgets and hardware scales. As illustrated in Fig.~\ref{fig:subarray}, both subarray settings are visualized on the full 12$\times$16 antenna layout, allowing a direct comparison of the sensing coverage and aperture reduction induced by each configuration.

\textbf{Subject-Specific Sub-benchmark.}
Human body shape, size, motion style, and habitual posture can vary substantially across subjects, which makes mesh reconstruction prone to identity-specific overfitting. This sub-benchmark can provide a stricter and more realistic evaluation of whether a model can capture subject-invariant structure rather than memorizing subject-dependent cues. This sub-benchmark therefore attempts to evaluate: \textbf{\textit{how well do mmWave-based HMR methods generalize to unseen subjects?}}

\subsection{Comparison Methods}
We compare against four baselines. 
(1) \textbf{mmMesh}\cite{mmmesh} takes mmWave point cloud sequences as input and combines a global
branch with a local (anchor-based) aggregation branch, followed by temporal modeling, to regress mesh-related and joint-related
outputs. 
(2) \textbf{M4esh}\cite{m4esh} also uses
point cloud sequences, where grouped local aggregation and temporal modeling are trained with joint supervision on
coarse/refined predictions to stabilize mesh reconstruction. 
(3) \textbf{P4Transformer}\cite{mmbody} is a temporal model (with spatio-temporal point modeling modules) that
regresses body parameters and reconstructs meshes/joints from mmWave point cloud sequences. 
(4) \textbf{Pc\_only (ours, ablation)} removes the tube encoder
and the cross-modal fusion from mmPTM, while keeping the same point cloud encoder, causal temporal aggregation and SMPL-X query decoder; thus it only consumes $X_{\mathrm{pc}}$ and isolates the gain from
tube cues and cross-modal fusion.

\subsection{Experiments}

To comprehensively evaluate the proposed HMR method and baselines under different generalization settings, five groups of experiments are conducted in this study.

\subsubsection{Base Setting}
This experimental setup is performed using data from all subjects and all actions. For each sequence with a length of $T=1000$ frames, the first 800 frames are used for training and the remaining 200 frames are used for testing. 

\subsubsection{Cross-subject Evaluation}
This experimental setup is conducted using data from all actions and all sequences. Among the 15 recruited subjects, 80\% of the subjects are used for training and the remaining 20\% are used for testing. 

\subsubsection{Cross-human-position Evaluation}
This experimental setup is conducted using data from all subjects. For the three sequences in Action Category I that correspond to different human positions, a three-fold cross-validation strategy is adopted. In each fold, two human-position settings are used for training and the remaining one is used for testing.

\subsubsection{Cross-human-orientation Evaluation}
This experimental setup is conducted using data from all subjects. For the three sequences in Action Category II that correspond to different human orientations, a three-fold cross-validation strategy is adopted.

\subsubsection{Subarray Size Evaluation}
This experimental setup is basically the same as the Base Setting, apart from the fact that point clouds generated from three array configurations, i.e., $3 \times 4$, $6 \times 8$, and $12 \times 16$, are used for comparison. To better isolate the influence of array size on reconstruction performance, the Pc\_only branch of our method is used in this experiment. Note that, except for this subarray size evaluation, the point clouds used in all other experiments are generated from the full $12 \times 16$ array.

\subsection{Training Details}
For mmPTM, we use a sliding window of length $L=5$. The point-cloud input is
$X_{\mathrm{pc}}\in\mathbb{R}^{L\times 256\times 6}$ with $N_{\mathrm{pc}}=256$ points per frame and feature order
$(x,y,z,v_r,E_{\mathrm{dB}},R)$, and the tube input is
$X_{\mathrm{tb}}\in\mathbb{R}^{L\times 40\times 68\times 40}$.
We optimize with AdamW using a constant learning rate of $1\times 10^{-4}$, weight decay $1\times 10^{-2}$, and
$(\beta_1,\beta_2)=(0.9,0.999)$ for $100$ epochs with batch size $64$ and gradient clipping at $5.0$.
AMP mixed-precision training is enabled, and experiments are conducted on NVIDIA RTX 3090 GPU. Unless otherwise stated, all methods are trained under a unified strategy for fair comparison.

Key architecture hyperparameters of mmPTM are fixed as follows: token dimension $d=128$; in the point cloud encoder, we use
$k=16$ neighbors with two local blocks and two set-abstraction stages that downsample
$256\!\rightarrow\!64\!\rightarrow\!32$ (radii $0.20/0.40$ with $32$ samples per neighborhood); the tube encoder is a
4-stage 3D CNN (base channels $=8$) with coordinate-bias injection and FiLM at each stage; cross-attention
uses $8$ heads (FFN dimension $512$); causal temporal aggregation uses a Transformer encoder with $5$ layers, $8$ heads,
FFN dimension $512$, and dropout $0.1$ with an upper-triangular causal mask; the SMPL-X query decoder uses $J=22$ pose
queries plus one translation query and one shape query.

\subsection{Evaluation Metrics}
We evaluate reconstruction quality from joint accuracy, mesh accuracy, rotation consistency, and global translation.

\subsubsection{Mean Per-Joint Position Error (MPJPE)}
MPJPE measures the average Euclidean distance between predicted and ground-truth 3D joints:
\begin{equation}
\mathrm{MPJPE}=\frac{1}{J}\sum_{j=0}^{J-1}\left\lVert \hat{J}_j-J_j\right\rVert_2,
\end{equation}
where $\hat{J}_j,J_j\in\mathbb{R}^{3}$ denote the predicted and ground-truth coordinates of the $j$-th joint.

\subsubsection{Mean Vertex Error (MVE)}
MVE measures the average Euclidean distance between predicted and ground-truth mesh vertices:
\begin{equation}
\mathrm{MVE}=\frac{1}{|\mathcal{V}|}\sum_{i\in\mathcal{V}}\left\lVert \hat{V}_i-V_i\right\rVert_2,
\end{equation}
where $\hat{V}_i,V_i\in\mathbb{R}^{3}$ are the $i$-th vertex coordinates.

\subsubsection{Mean Rotation Error (MRE)}
MRE evaluates rotation discrepancies on $SO(3)$ using the geodesic angle:
\begin{equation}
\mathrm{MRE}=\frac{1}{J}\sum_{j=0}^{J-1}\theta\!\left(\hat{\mathbf{R}}_j^{\top}\mathbf{R}_j\right),
\end{equation}
where $\hat{\mathbf{R}}_j,\mathbf{R}_j\in SO(3)$ are predicted and ground-truth rotation matrices, and
$\theta(\cdot)$ denotes the minimum rotation angle of the relative rotation.

\subsubsection{Mean Location Error (MLE)}
MLE measures the Euclidean error of the global translation on the ground-plane:
\begin{equation}
\mathrm{MLE}=\left\lVert \hat{\mathbf{t}}_{xz}-\mathbf{t}_{xz}\right\rVert_2,\qquad
\mathbf{t}_{xz}=(t_x,t_z)\in\mathbb{R}^{2},
\end{equation}
where $\hat{\mathbf{t}},\mathbf{t}\in\mathbb{R}^{3}$ are the predicted and ground-truth translations.

\section{Result and Analysis}

In this section, we present a comprehensive analysis of the proposed benchmark under different evaluation settings. Beyond the quantitative results reported in Tables~\ref{tab:base_cross_subject}--\ref{tab:subarray_size}, we also provide qualitative visualization and analyze motions with and without large global displacement. These results collectively reveal that our proposed method consistently achieves the strongest accuracy and generalization capability in most settings, while the proposed benchmark exposes clear generalization gaps caused by changes in subject identity, human position, human orientation, and array aperture. More importantly, the observed trends indicate that mmWave-based HMR is highly sensitive to measurement configuration shifts, and that improved sensing coverage is essential for robust mesh reconstruction.

\begin{table*}[t]
\centering
\caption{Results on the base setting and cross-subject evaluation. Lower is better for all metrics.}
\label{tab:base_cross_subject}
\begin{tblr}{
  cells = {c},
  cell{1}{1} = {r=2}{},
  cell{1}{2} = {c=4}{},
  cell{1}{6} = {c=4}{},
  vline{2,6} = {1-Z}{0.03em},
  hline{1,Z} = {-}{0.1em},
  hline{3} = {-}{0.03em},
}
Method          & Base Setting &       &         &       & Cross-subject Evaluation &       &        &       \\
                & MPJPE~       & MVE~  & MRE~    & MLE~  & MPJPE~        & MVE~  & MRE~   & MLE~  \\
mmMesh\cite{mmmesh}          & 2.94        & 3.37 & 7.22   & 1.90 & 5.88         & 6.74 & 11.80  & 3.11 \\
M4esh\cite{m4esh}           & 3.27        & 4.76 & 107.74 & 1.90 & 5.17         & 6.24 & 54.75 & \textbf{2.45} \\
P4Transformer\cite{mmbody}   & 3.06        & 3.52 & 5.44   & 2.02 & 5.42         & 6.35 & 7.35  & 2.69 \\
Pc\_only (ours) & 2.58        & 2.84 & 3.12   & 1.73 & 5.16         & 5.72 & 6.81  & 2.61 \\
mmPTM (ours)    & \textbf{2.34}        & \textbf{2.59} & \textbf{2.89}   & \textbf{1.52} & \textbf{5.07}         & \textbf{5.60} & \textbf{6.62}  & 2.50 
\end{tblr}

\vspace{2pt}
\begin{minipage}{\linewidth}
\raggedright\scriptsize
\emph{Note:} For all results in this paper, MPJPE, MVE, and MLE are reported in centimeters (cm), while MRE is reported in degrees ($^\circ$).
\end{minipage}
\end{table*}
\begin{figure*}[t]

  \centering
  
  \includegraphics[width=0.8\textwidth]{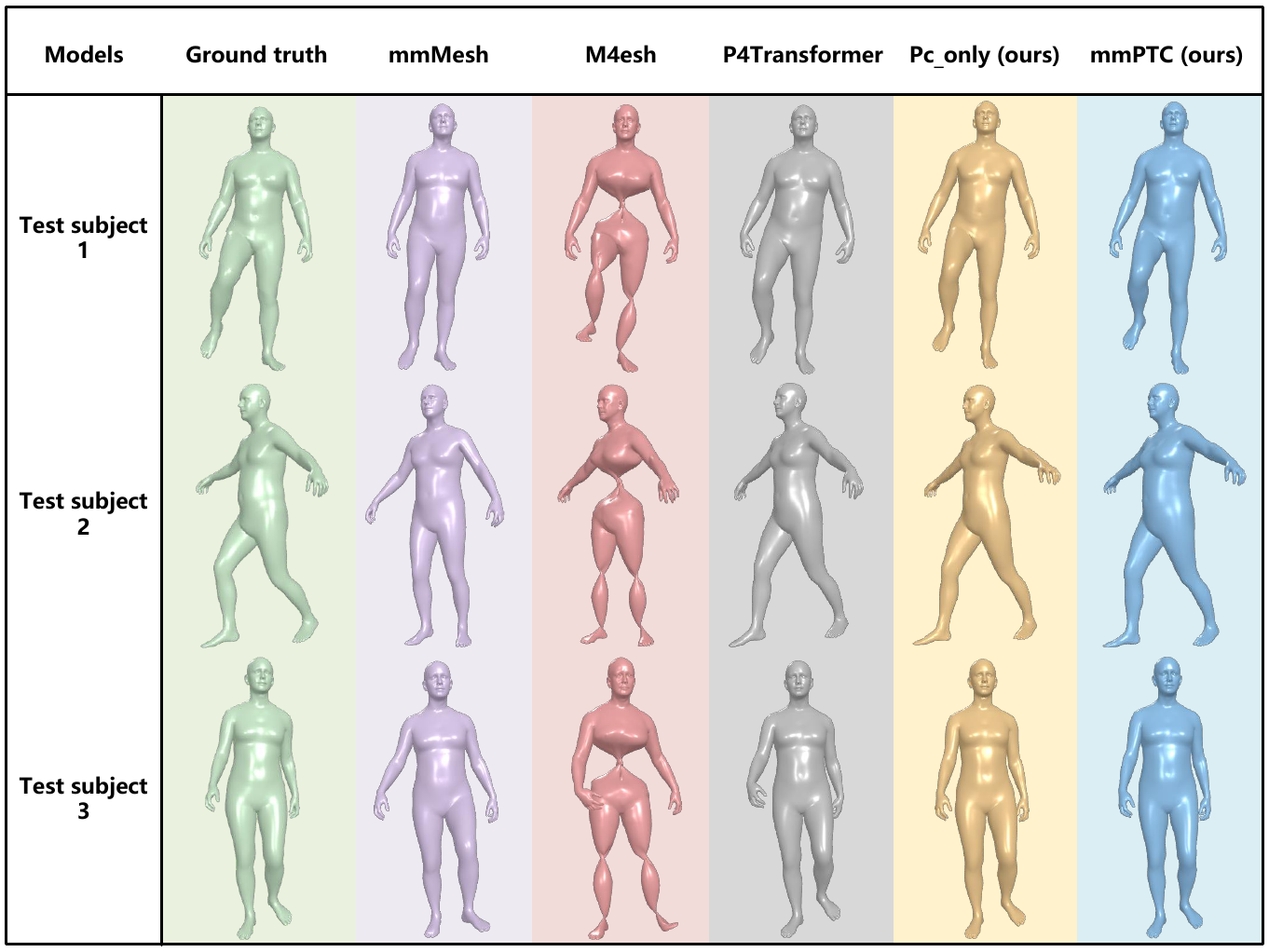}
  
 \caption{Visualization of cross-subject evaluation results. The first column shows the ground-truth meshes, while the remaining columns present the meshes reconstructed by the corresponding models on test data from different subjects.}

  \label{fig:result}
  
\end{figure*}

\subsection{Overall Performance on the Base Setting and Cross-Subject Evaluation}

As shown in Table~\ref{tab:base_cross_subject}, mmPTM achieves the best performance on nearly all metrics under both the base setting and cross-subject evaluation, while Pc\_only consistently ranks second and outperforms the other baselines. Under the base setting, where the training and testing data share the same subjects and measurement conditions, the superior performance of mmPTM indicates that the proposed multi-radar representations, together with the designed reconstruction strategy, can more effectively model human geometric structures and body surface details.

The cross-subject evaluation is more challenging, as the test subjects exhibit different body shapes, body sizes, motion habits, and local shape characteristics from those seen during training. To better visualize the experimental results, Fig.~\ref{fig:result} presents representative reconstruction results under cross-subject evaluation, where the first column shows the ground-truth meshes and the remaining columns show the reconstructed meshes from different methods. It can be observed that M4esh produces noticeably distorted mesh shapes in several cases, which is consistent with its relatively high MRE reported in Table~\ref{tab:base_cross_subject}. This behavior can be attributed to the fact that M4esh is trained with coarse/fine skeletal alignment and heavily weighted joint/vertex supervision, which may overemphasize surface and joint fitting while underfitting the rotational structure of the human body. As a result, the method can still achieve reasonably good geometric alignment, but tends to produce larger articulated pose errors. This also highlights an important distinction between the evaluation metrics: MPJPE and MVE mainly measure direct surface-level reconstruction accuracy, i.e., how accurately the recovered joints and vertices align with the ground-truth mesh, whereas MRE reflects articulated pose correctness and kinematic consistency.

Overall, both the visualized results and Table~\ref{tab:base_cross_subject} show that all methods experience performance degradation under cross-subject evaluation compared with the base setting, which is expected because subject identity changes introduce a more challenging generalization analysis problem. Nevertheless, mmPTM remains the strongest method across the two settings, suggesting that the proposed multi-radar representations and reconstruction strategy are able to learn more robust identity-invariant representations than the competing baselines.

\noindent\textbf{Take-away Message:}  
These results suggest two important findings. First, for HMR, articulated pose constraints are crucial; overemphasizing purely geometric alignment metrics such as MPJPE and MVE may lead to visually distorted reconstructions and worse pose recovery. Second, all methods degrade under cross-subject evaluation relative to the base setting, and no clear outliers are observed, confirming that subject variation is a meaningful factor for generalization analysis in mmWave-based HMR.

\begin{table*}[t]
\centering
\caption{Results on the cross-human-position evaluation. Lower is better for all metrics.}
\label{tab:cross_radar}
\begin{tblr}{
  cells = {c},
  cell{1}{1} = {r=2}{},
  cell{1}{2} = {c=4}{},
  cell{1}{6} = {c=4}{},
  cell{1}{10} = {c=4}{},
  vline{2,6,10} = {1-Z}{0.03em},
  hline{1,Z} = {-}{0.1em},
  hline{3} = {-}{0.03em},
}
Method          & Position 1,2$\rightarrow$3 &        &       &        & Position 1,3$\rightarrow$2 &       &       &       & Position 2,3$\rightarrow$1 &        &        &        \\
                & MPJPE~         & MVE~   & MRE~  & MLE~   & MPJPE~         & MVE~  & MRE~  & MLE~  & MPJPE~         & MVE~   & MRE~   & MLE~   \\
mmMesh\cite{mmmesh}          & 117.53         & 117.36 & 12.35 & 116.23 & 57.84          & 58.02 & 54.60 & 56.15 & 54.78          & 54.85  & 64.64  & 52.72  \\
M4esh\cite{m4esh}           & 108.96         & 115.24 & 90.36 & 107.05 & 56.43          & 61.50 & 58.00 & 54.37 & 35.18          & 37.69  & 77.77  & 28.27  \\
P4Transformer\cite{mmbody}   & 107.99         & 104.54 & 15.65 & 106.68 & 50.48          & 47.02 & 13.05  & 48.18 & 23.65          & 23.79  & 12.75   & 21.73  \\
Pc\_only (ours) & 112.49         & 111.19 & \textbf{10.05}  & 111.62 & 54.53          & 52.11 & 9.26  & 53.86 & 16.80          & 18.28  & 9.11   & 12.98  \\
mmPTM (ours)    & \textbf{93.68}          & \textbf{91.03}  & 10.48  & \textbf{92.91}  & \textbf{41.69}          & \textbf{41.91} & \textbf{8.64}  & \textbf{40.57} & \textbf{10.04}          & \textbf{11.76}  & \textbf{8.35}   & \textbf{6.04}   
\end{tblr}
\end{table*}

\subsection{Generalization under Human Position Shifts}

Table~\ref{tab:cross_radar} reports the results of the cross-human-position evaluation under three-fold cross-validation, namely Position~1,2$\rightarrow$3, Position~1,3$\rightarrow$2, and Position~2,3$\rightarrow$1. The specific location information of these three settings has been explained in \textit{Human Position Variation Sub-benchmark}. As shown in the table, mmPTM achieves the best overall performance in all three folds. However, the absolute errors are still substantially larger than those observed in the base setting, indicating that human position shifts remain a challenging source of domain discrepancy for HMR. 

To avoid being dominated by occasional MRE outliers caused by insufficient pose supervision in some baseline methods, we first analyze the overall trend using the remaining metrics and then discuss MRE separately. A clear trend can be observed across the three folds from all methods. In Position~1,2$\rightarrow$3, where the test position is far away from the two training positions, the error level becomes the highest, with mmPTM still performing best but the overall error remaining at around 90~cm, while the competing baselines are around 110~cm. In Position~1,3$\rightarrow$2, the test position is closer to the training coverage, and the overall error is reduced to around 40~cm for mmPTM and around 50~cm for the baselines. More interestingly, in Position~2,3$\rightarrow$1, the test position lies between the two training positions in the position space, and mmPTM reduces the overall error to below 10~cm, which is within an acceptable range for this situation, whereas the baseline methods still remain above 20~cm. This pattern suggests that the generalization difficulty is strongly governed by whether the test position lies inside or outside the region spanned by the training positions. In other words, human-position generalization behaves more like a distribution-support problem than a simple sensor-geometry perturbation.

Regarding MRE, mmPTM and Pc\_only still maintain relatively low values under the harder folds, which suggests that the predicted articulated structure does not collapse even when absolute alignment becomes difficult. This indicates that the learned representations retain a meaningful prior over human pose, and that the proposed model is able to preserve coarse kinematic consistency under severe position shifts.

\noindent\textbf{Take-away Message:}  
Human-position generalization is strongly influenced by the distribution coverage in the sensing space. When the test position lies outside the spatial support spanned by the training positions, all methods degrade significantly; when it lies within that support, the reconstruction error decreases markedly. This further confirms that the proposed benchmark is effective for generalization analysis under human-position shifts, and that expanding the spatial coverage of human body placements in the training set is important for improving mmWave-based HMR performance.

\begin{table*}[t]
\centering
\caption{Results on the cross-human-orientation evaluation. Lower is better for all metrics.}
\label{tab:cross_orientation}
\begin{tblr}{
  cells = {c},
  cell{1}{1} = {r=2}{},
  cell{1}{2} = {c=4}{},
  cell{1}{6} = {c=4}{},
  cell{1}{10} = {c=4}{},
  vline{2,6,10} = {1-Z}{0.03em},
  hline{1,Z} = {-}{0.1em},
  hline{3} = {-}{0.03em},
}
Method          & Orientation 1,2$\rightarrow$3 &       &       &      & Orientation 1,3$\rightarrow$2 &       &        &      & Orientation 2,3$\rightarrow$1 &        &        &        \\
                & MPJPE~         & MVE~  & MRE~  & MLE~ & MPJPE~         & MVE~  & MRE~   & MLE~ & MPJPE~         & MVE~   & MRE~   & MLE~   \\
mmMesh\cite{mmmesh}          & 28.45          & 34.94 & 78.92 & 10.32 & 21.97          & 26.99 & 63.29  & 8.97 & 18.10          & 21.73  & 20.11  & 10.00  \\
M4esh\cite{m4esh}           & 27.19          & \textbf{34.20} & 103.99 & \textbf{6.81} & \textbf{18.63}          & \textbf{23.36} & 131.60 & 6.86 & 16.72          & 21.00  & 99.16  & 6.59   \\
P4Transformer\cite{mmbody}   & 26.77          & 34.83 & 50.23 & 9.78 & 23.97          & 29.55 & 72.95  & 9.24 & 16.67          & 20.29  & 15.44  & 7.29   \\
Pc\_only (ours) & \textbf{26.68}          & 34.61 & 17.54 & 7.15 & 24.05          & 30.36 & 13.70  & \textbf{5.64} & 15.32          & 18.40  & 14.03  & 6.77   \\
mmPTM (ours)    & 26.82          & 34.81 & \textbf{16.98} & 7.60 & 22.42          & 26.43 & \textbf{13.32}  & 6.10 & \textbf{13.67}          & \textbf{16.22}  & \textbf{12.83}  & \textbf{6.56}   
\end{tblr}
\end{table*}

\subsection{Generalization under Human Orientation Shifts}

Table~\ref{tab:cross_orientation} reports the results of the cross-human-orientation evaluation under three-fold cross-validation, namely Orientation~1,2$\rightarrow$3, Orientation~1,3$\rightarrow$2, and Orientation~2,3$\rightarrow$1. The specific orientation information of these three settings has been explained in \textit{Human Orientation Variation Sub-Benchmark}. Similarly to the analytical method and observed trend of the cross-human-position evaluation, it can be seen that the Orientation~2,3$\rightarrow$1 setting generally yields lower reconstruction error than Orientation~1,2$\rightarrow$3 and Orientation~1,3$\rightarrow$2. This suggests that the coverage of human orientations in the training data plays an important role in determining the generalization performance of mmWave-based HMR. In particular, when the test orientation lies within or near the orientation support spanned by the training data, the model can better adapt to the resulting changes in radar observations; otherwise, the reconstruction error increases accordingly. However, this trend is less pronounced than that in the cross-human-position evaluation, indicating that human-orientation shifts also affect HMR generalization, but to a milder extent. This is reasonable because changing the human orientation primarily alters the radar-facing surface, backscatter distribution, and self-occlusion pattern, while the absolute subject-radar displacement remains largely unchanged. Overall, these results indicate that mmWave-based HMR is sensitive to body-orientation shifts, while also showing that mmPTM provides more stable performance under such orientation-dependent changes.

\noindent\textbf{Take-away Message:}  
Human-orientation generalization is influenced by the coverage of body orientations in the training data, but the effect is less severe than that of human-position shifts. When the test orientation is closer to the orientation support of the training data, reconstruction performance improves accordingly. This confirms that the proposed benchmark is effective for generalization analysis under orientation shifts, and that collecting sufficiently diverse orientation patterns is also important for practical mmWave-based HMR.

\begin{table*}[t]
\centering
\caption{Results on the subarray size evaluation. Lower is better for all metrics.}
\label{tab:subarray_size}
\begin{tblr}{
  cells = {c},
  cell{1}{1} = {r=2}{},
  cell{1}{2} = {c=4}{},
  cell{1}{6} = {c=4}{},
  cell{1}{10} = {c=4}{},
  vline{2,6,10} = {1-Z}{0.03em},
  hline{1,Z} = {-}{0.1em},
  hline{3,7} = {-}{0.03em},
}
Method          & 3$\times$4    &      &       &      & 6$\times$8    &      &       &      & 12$\times$16  &      &        &      \\
                & MPJPE~ & MVE~ & MRE~  & MLE~ & MPJPE~ & MVE~ & MRE~  & MLE~ & MPJPE~ & MVE~ & MRE~   & MLE~ \\
mmMesh\cite{mmmesh}          & 5.13   & 5.81 & 9.98  & 3.15 & 4.30   & 4.92 & 8.67  & 2.63 & 2.94   & 3.37 & 7.22   & 1.90 \\
M4esh\cite{m4esh}           & 4.20   & 5.05 & 26.56 & 2.38 & 3.67   & 4.86 & 79.66 & 2.12 & 3.27   & 4.76 & 107.74 & 1.90 \\
P4Transformer\cite{mmbody}   & 3.82   & 4.45 & 5.86  & 2.27 & 3.18   & 3.70 & 5.53  & 1.96 & 3.06   & 3.52 & 5.44   & 2.02 \\
Pc\_only (ours) & \textbf{3.31}   & \textbf{3.66} & \textbf{3.67}  & \textbf{2.13} & \textbf{2.82}   & \textbf{3.12} & \textbf{3.29}  & \textbf{1.81} & \textbf{2.58}   & \textbf{2.84} & \textbf{3.12}   & \textbf{1.73} \\
Pc\_only (Action Category I) & 2.25 & 2.61 & 2.78 & 1.49 & 1.95 & 2.27 & 2.57 & 1.22 & 1.82 & 2.11 & 2.49 & 1.22 \\
Pc\_only (Action Category III) & 6.34 & 6.94 & 5.89 & 3.86 & 5.27 & 5.72 & 5.22 & 3.37 & 4.67 & 4.99 & 4.83 & 3.09 \\
\end{tblr}
\end{table*}

\subsection{Reconstruction Performance under Varying Array Apertures}

Table~\ref{tab:subarray_size} reports the results under three progressively larger subarray configurations, namely $3 \times 4$, $6 \times 8$, and $12 \times 16$. As expected, the mesh reconstruction performance of most methods improves consistently as the array aperture increases. This trend is consistent with radar sensing principles: a larger array aperture provides finer angular resolution, stronger spatial discrimination capability, and more informative representations. In contrast, a smaller subarray offers fewer spatial degrees of freedom and therefore makes accurate mesh reconstruction more difficult. 

A more detailed pattern can be observed by examining the last two rows of Table~\ref{tab:subarray_size}, which correspond to Pc\_only evaluated on Action Category~I and Action Category~III, respectively. Recall that Action Category~I contains actions with a fixed global position, whereas Action Category~III corresponds to free walking, where the subject traverses a larger spatial range. As shown in the table, Pc\_only on Action Category~III exhibits much larger errors than on Action Category~I across all subarray sizes, indicating that global displacement and larger motion extent make reconstruction substantially more challenging. At the same time, the performance gain brought by increasing the aperture from $3 \times 4$ to $6 \times 8$ and $12 \times 16$ is more pronounced for Action Category~III than for Action Category~I. This suggests that larger subarrays are especially beneficial when the motion induces stronger spatial variation in the radar observations. From a practical perspective, these results also indicate that algorithm design and hardware aperture should be considered jointly: a stronger reconstruction model can recover part of the performance sacrificed by using a more compact and portable radar configuration, while the benefit of a larger aperture is most evident for motions with larger global displacement.

\noindent\textbf{Take-away Message:}  
Larger array apertures generally lead to better HMR performance, but the extent of improvement depends on the motion pattern. Motions with larger global displacement, such as walking, benefit more from increased aperture than motions with fixed global position. These results confirm that the proposed benchmark is effective for studying the interaction between sensing resolution and motion complexity.

\section{Conclusion}

This study introduces DGHMesh,  a large-scale dual-radar mmWave dataset and generalization-focused benchmark for HMR. DGHMesh provides synchronized FMCW and SFCW radar data, RGB images, precise 3D human mesh annotations, and accurately calibrated radar spatial positions, together with diverse measurement configurations covering human position, human orientation, and subarray size. To demonstrate its effectiveness, we propose mmPTM, a multi-radar fusion network that jointly leverages point cloud and imaging tube representations. Extensive experiments are conducted with mmPTM and several representative baselines across multiple sub-benchmarks. The results show that mmPTM consistently achieves outstanding accuracy and competitive generalization capability under diverse conditions. We expect DGHMesh to provide a valuable testbed for the community, enabling systematic evaluation of mmWave-based HMR methods and promoting future research on multi-modal fusion, robust HMR, generalization analysis, and practical mmWave-based human sensing systems.

\bibliographystyle{ieeetr} 
\bibliography{references} 

\end{document}